\newif\ifdraft \draftfalse
\newif\ifarxiv \arxivtrue

\ifarxiv
\documentclass[acmtog,balance=false,nonacm,authorversion]{acmart}
\else
\documentclass[acmtog,balance=false,authorversion]{acmart}
\fi
\ifarxiv
\else
\fi
\AtBeginDocument{%
  }

\copyrightyear{2026}
\acmYear{2026}
\ifarxiv
\setcopyright{none}
\acmDOI{}
\else
  \setcopyright{none}
  \renewcommand\footnotetextcopyrightpermission[1]{}
  \acmConference[SIGGRAPH Conference Papers '26]{Special Interest Group on Computer Graphics and Interactive Techniques Conference Conference Papers}{July 19--23, 2026}{Los Angeles, CA, USA}
  \acmBooktitle{Special Interest Group on Computer Graphics and Interactive Techniques Conference Conference Papers (SIGGRAPH Conference Papers '26), July 19--23, 2026, Los Angeles, CA, USA}
  \acmDOI{10.1145/3799902.3811092}
  \acmISBN{979-8-4007-2554-8/2026/07}
  \acmVolume{XX}
  \acmNumber{XX}
  \acmArticle{XX}
  \acmMonth{8}
  \acmSubmissionID{papers 482}
\fi

\usepackage{multirow}
\usepackage{xspace}
\usepackage{cleveref}
\usepackage{algorithm}
\usepackage{algorithmic}
\usepackage{stfloats}

\newif\ifappendix
\appendixfalse
\newcommand{\appsecref}[2]{\ifappendix Appendix~\ref{#1}\else sup. mat. Sec.~#2\fi}

\DeclareMathOperator*{\argmin}{arg\,min}

\definecolor{PositiveColor}{rgb}{.7,1,.7}
\definecolor{NegativeColor}{rgb}{1,.7,.7}
\definecolor{darkblue}{rgb}{0.2,0.4,0.6}
\definecolor{darkgreen}{rgb}{0.2,0.5,0.2}

\crefname{section}{Sec.}{Secs.}
\Crefname{section}{Section}{Sections}
\Crefname{table}{Table}{Tables}
\crefname{table}{Tab.}{Tabs.}
\Crefname{figure}{Figure}{Figures}
\crefname{figure}{Fig.}{Figs.}
\Crefname{algorithm}{Algorithm}{Algorithms}
\crefname{algorithm}{Alg}{Algs}

\newcommand{\algoname}{MultiAct\xspace}
 
\newcommand{\schemename}{ParamGate\xspace} %

\newcommand{\R}{\mathbb{R}} %
\newcommand{\E}{\mathbb{E}} %

\newcommand{\Dcal}{\mathcal{D}}

\newcommand{\Ecal}{\mathcal{E}}

\newcommand{\Ical}{\mathcal{I}}
\newcommand{\Ncal}{\mathcal{N}}
\newcommand{\Mcal}{\mathcal{M}}

\newcommand{\Lcal}{\mathcal{L}}

\newcommand{\Scal}{\mathcal{S}}
\newcommand{\Tcal}{\mathcal{T}}
\newcommand{\Ycal}{\mathcal{Y}}
\newcommand{\Loss}{\Lcal}

\newcommand{\abs}[1]{\lvert #1 \rvert}
\newcommand{\norm}[1]{\lVert#1\rVert}

\ifdraft
\newcommand{\AAAA}[1]{{\color{red}\textbf{Andreas: }#1}\normalfont}

\newcommand{\ofir}[1]{{\color{magenta}\textbf{Ofir: }#1}\normalfont}
\newcommand{\review}[1]{{\color{blue}#1}\normalfont}
\newcommand{\srcolor}{violet}
\newcommand{\src}[1]{\textcolor{\srcolor}{SR: #1}}

\newcommand{\nscolor}{blue}
\newcommand{\nat}[1]{{\color{\nscolor}\textbf{Nathan: }#1}\normalfont}
\newcommand{\nsc}[1]{\textcolor{\nscolor}\textbf{{NS:} #1}}

\newcommand{\dcc}[1]{{\color{red}\textbf{DC:} #1}}

\newcommand{\ascolor}{darkgreen}
\newcommand{\asc}[1]{{\color{\ascolor}\textbf{AS:} #1}}

\newcommand{\todo}[1]{{\color{red} [#1]}}
\else
\newcommand{\AAAA}[1]{}

\newcommand{\ofir}[1]{}
\newcommand{\review}[1]{}
\newcommand{\src}[1]{}

\newcommand{\nat}[1]{}
\newcommand{\nsc}[1]{}

\newcommand{\dcc}[1]{}

\newcommand{\asc}[1]{}

\newcommand{\todo}[1]{}
\fi

\newcommand{\eg}{\emph{e.g.}\xspace}

\newcommand{\ie}{\emph{i.e.}\xspace}

\appendixtrue

\begin{document}
\title[\algoname]{\algoname: Text-to-Motion Generation from Composite Text via Tailored Attention Guidance}

\author{Nathan Sala}
\affiliation{%
  \institution{Tel Aviv University}
  \city{Tel Aviv}
  \postcode{4610101}
  \country{Israel}}
\email{natouda@gmail.com}

\author{Ofir Abramovich}
\orcid{0009-0007-4248-6423}
\affiliation{%
  \institution{Reichman University}
  \city{Herzliya}
  \postcode{4610101}
  \country{Israel}}
\affiliation{%
  \institution{CYENS Centre of Excellence}
  \city{Nicosia}
  \postcode{1016}
  \country{Cyprus}}
\email{ofir1080@gmail.com}

\author{Ariel Shamir}
\orcid{0000-0001-7082-7845}
\affiliation{%
  \institution{Reichman University}
  \city{Herzliya}
  \country{Israel}}
\email{arik@runi.ac.il}

\author{Daniel Cohen-Or}
\orcid{0000-0001-6777-7445}
\affiliation{%
  \institution{Tel Aviv University}
  \city{Tel Aviv}
  \country{Israel}}
\email{cohenor@gmail.com}

\author{Andreas Aristidou}
\orcid{0000-0001-7754-0791}
\affiliation{
  \institution{University of Cyprus}
  \city{Nicosia}
  \postcode{1678}
  \country{Cyprus}}
\affiliation{%
  \institution{CYENS Centre of Excellence}
  \city{Nicosia}
  \postcode{1016}
  \country{Cyprus}}
\email{a.aristidou@ieee.org}

\author{Sigal Raab}
\orcid{0000-0001-6616-257X}
\affiliation{%
  \institution{Tel Aviv University}
  \city{Tel Aviv}
  \country{Israel}}
\email{sigal.raab@gmail.com}

\renewcommand\shortauthors{Sala et al.}

\ifarxiv
\else
\ccsdesc[500]{Computing methodologies~Computer graphics}
\ccsdesc[300]{Computing methodologies~Artificial intelligence}
\fi

\begin{abstract}

Text-to-motion generation has progressed rapidly in recent years, offering an expressive interface for animation and human–computer interaction. However, current models remain brittle when handling prompts that describe multiple actions occurring at the same time. Rather than realizing all components of a composite description, models frequently prioritize a single dominant action and neglect the rest, leading to incomplete or ambiguous motion. 
We present \algoname, an unpaired, inference-time framework for compositional text-to-motion synthesis that operates directly on pretrained motion generators without retraining or architectural modification. Our method counteracts semantic collapse by adaptively amplifying cross-attention scores associated with underrepresented prompt components.
We note that effective modulation depends on prompt-specific choices, such as which tokens and layers to target, and introduce a lightweight auxiliary 
decision scheme that determines the most effective attention-strengthening parametrization.
Extensive quantitative and qualitative evaluations demonstrate that \algoname consistently outperforms existing baselines on composite prompts, achieving improved semantic coverage while preserving motion realism.
Project page: \url{https://natsala13.github.io/multiact.github.io}.

\end{abstract}

\begin{teaserfigure}
    \centering
    \includegraphics[width=\textwidth]{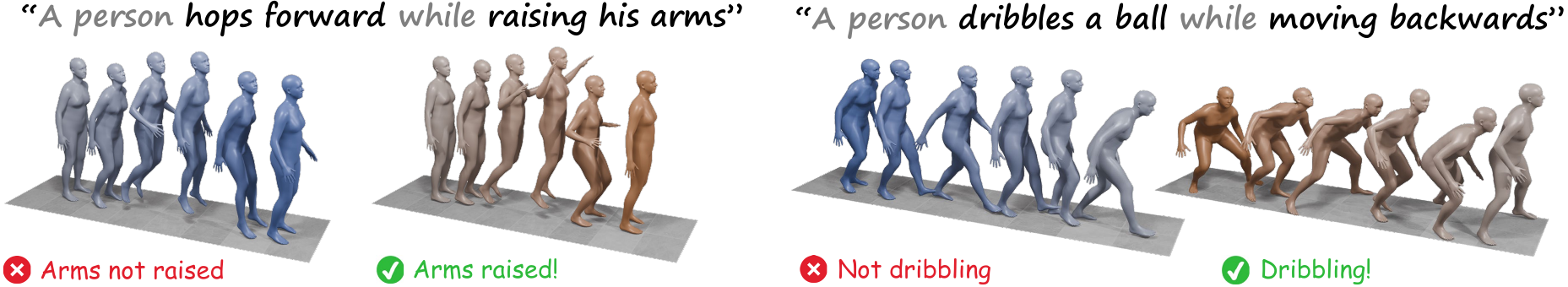}
    \caption{\algoname synthesizes motion from composite textual descriptions by selectively modulating cross-attention to amplify weakly represented elements in the prompt. \textbf{Blue:} Backbone text-to-motion synthesis fails to generate key action components, such as raising the arms while hopping forward (left) and dribbling a ball while moving backward (right). \textbf{Brown:} Our framework successfully generates all action primitives specified in the prompt. %
    Color saturation indicates time progression; higher saturation indicates later times.}
    \label{fig:teaser}
    \Description[Two example composite prompts comparing the backbone and MultiAct]{
    The teaser shows two side-by-side text-to-motion examples. In each example, the backbone misses one action component while MultiAct generates a motion that satisfies both requested actions.}
\end{teaserfigure}

\maketitle

\section{Introduction}
\label{sec:Introduction}

Human motion synthesis aims to generate realistic, physically plausible motion while respecting high-level semantic intent. Early data-driven methods focused on kinematic realism or action-specific generation~\cite{holden2016deep,petrovich2021actor,Guo:2020}. Subsequent models have substantially expanded this paradigm by enabling motion synthesis from abstract controls such as natural language~\cite{Tevet:2023:MDM, Zhang:2024, Guo:2022, guo2024momask}. These text-to-motion models thus provide an intuitive interface for animation, virtual agents, and human–computer interaction.

Despite this progress, most text-to-motion models excel only for \emph{single atomic actions} and struggle with composite prompts. For complex descriptions, they often attend primarily to the dominant verb and ignore secondary constraints, a phenomenon we call \emph{vanishing semantics}. As a result, the generated motion only partially satisfies the prompt, failing to realize all specified actions and details (\eg, adverbs, direction, or manner). We attribute this to entangled cross-attention: when multiple semantic components compete, attention mass collapses onto a few dominant tokens, suppressing the rest.

In this work, we study the generation of \textit{simultaneous actions} from \emph{composite prompts}, such as “running while waving arms,” where multiple actions must co-occur in time and may engage overlapping sets of body parts. This setting differs fundamentally from \textit{sequential actions} (\eg, “walk, then jump”), which can often be handled via concatenation or temporal planning~\citep{shafir2024human,zhang2023diffcollage,athanasiou2022teach}. We consider a motion generation as successful when all semantic components are visibly present and co-occur for at least part of the sequence, without prescribing when overlap must occur. Successful synthesis of simultaneous actions requires preserving multiple constraints concurrently, without allowing any single action to dominate. 

Existing approaches to co-occurring actions rely on restrictive assumptions. For example, STMC~\citep{Petrovich:2024} 
requires structured inputs (explicit action tracks) and prior body–text correspondences, limiting general inference-time control.
Other methods rely on auxiliary components, such as a graph reasoning network \cite{jin2023act,chang2025casim}, which complicates the setup.

Inspired by inference-time attention manipulation in text-to-image generation~\citep{Chefer:2023,rassin2023Linguistic}, we explore similar control for motion. However, directly transferring these techniques is unstable: motion’s spatio-temporal nature makes temporal coherence highly sensitive to attention perturbations, and effective settings are prompt-dependent (\eg, token/layer choices).

We introduce \algoname, an unpaired, inference-time, optimization-based framework for compositional text-to-motion generation. \algoname operates on a pretrained motion backbone without fine-tuning or architectural changes, and mitigates vanishing semantics by selectively amplifying cross-attention for underrepresented semantics (defined in \cref{sec:attn_guiding}). This method relies on several parameters: the token to strengthen, the transformer layer to which it is applied, and the diffusion-step schedule. 
Parameter choice is prompt-dependent; therefore, we propose a lightweight parameter-selection scheme,  \schemename, which predicts these parameters, eliminating the need for exhaustive search and enabling generalization across prompts.

Together, these components enable reliable compositional control for simultaneous-action synthesis while preserving motion realism and temporal coherence, as illustrated in \cref{fig:teaser}. 
Quantitative and qualitative results show that \algoname consistently outperforms prior baselines, including ones specializing in composite-motion.

\begin{figure}[t]
    \centering
    \includegraphics[width=\linewidth]{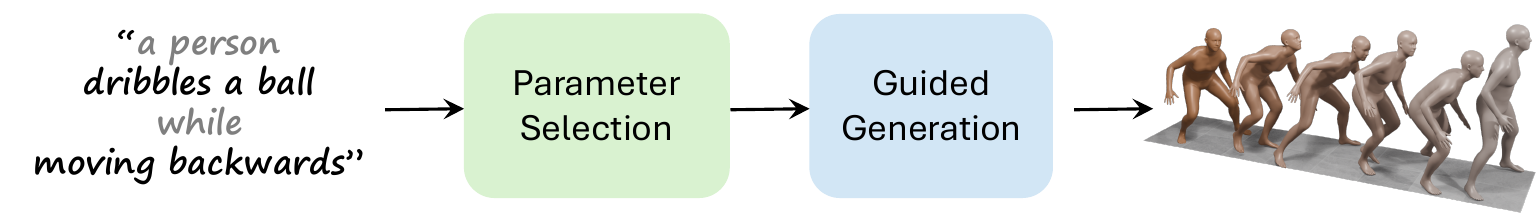}
    \caption{
        \textbf{Pipeline.} 
        Given a text prompt, \algoname selects prompt-specific parameters (\cref{sec:param_select}), applies tailored guided generation (\cref{sec:attn_guiding}), and outputs the resulting motion.
    }
    \label{fig:pipeline}
    \Description[Overview of the MultiAct pipeline]{
    The pipeline starts from a text prompt, selects prompt-specific guidance parameters, applies tailored attention-guided generation during diffusion inference, and outputs the final synthesized motion.}
\end{figure}

\section{Related Work}
\label{sec:Related_Work}

\paragraph{Text-conditioned human motion generation}
Text-to-motion generation has been widely studied using both discrete and continuous motion representations. Discrete approaches typically quantize continuous motion into codebook tokens and generate motion autoregressively using transformers conditioned on text \cite{pinyoanuntapong2024mmm, pinyoanuntapong2024bamm, pinyoanuntapong2025maskcontrol, guo2022tm2t, guo2024momask, guo2025snapmogen, javed2024intermask, jiang2024motiongpt,  zhang2023generating, zhang2025kinmo, qian2024text}. %
While this formulation simplifies training, it introduces quantization errors that degrade fine-grained motion details. Autoregressive generation, whether temporal or mask-based, is unidirectional and thus prone to error accumulation.

Continuous text-to-motion generation is dominated by diffusion-based models, which synthesize motion by iteratively denoising input noise conditioned on text \cite{zhang2024motion, Tevet:2023:MDM, qian2025physdiff}. Diffusion can operate in pose space or in learned latent spaces: pose-space diffusion \cite{raab2024single,tevet2025closd} is sensitive to mocap noise, whereas latent diffusion mitigates this by compressing motion via VAEs \cite{chen2023executing, dai2024motionlcm, meng2025absolute, zhang2025flashmo, zhang2024motion, Weng2026realign}. However, single-vector latents may obscure temporal detail, motivating structured or autoregressive latent diffusion methods \cite{meng2024rethinking, tu2025autoregressive, xiao2025motionstreamer, zhu2025motiongpt3}.

\paragraph{Composite text to motion}

Many text-to-motion models \cite{Tevet:2023:MDM,chen2023executing,zhang2023generating,huang2024controllable} use a single textual token to encode the entire prompt, based on CLIP \cite{radford2021learning} or Sentence BERT \cite{reimers2019sentence} encoding. Others \cite{tevet2025closd,guo2025snapmogen} use a more detailed encoding of the prompt, based on BERT \cite{devlin-etal-2019-bert} or T5 \cite{raffel2020exploring}. However, these works struggle when conditioned on prompts containing simultaneous actions.  

Some methods enrich text representations with finer-grained cues. CoMo \cite{huang2024controllable} and FGMDM \cite{shi2023generating} incorporate GPT-4–generated keywords, while FineMoGen \cite{zhang2023finemogen} use explicit spatial and temporal descriptions. GraphMotion \cite{jin2023act} and CASIM \cite{chang2025casim} convert text into semantic graphs processed by graph-based reasoning and coarse-to-fine diffusion. SINC \cite{Athanasiou:2023} and STMC \cite{Petrovich:2024} support co-occurring actions through timeline-level control by assigning text conditions to body parts, while GuidedMotion \cite{guidedMotion} construct an action tree to manipulate the diffused data.

\paragraph{Attention editing in generative models}
Attention control originated in image diffusion and was later extended to motion.
In the imaging domain, 
several works
\citep{tumanyan2022plugandplay, cao2023masactrl, alaluf2023crossimagema} show that the self-attention layers encode structural information that can be used to edit an image without losing its original composition.
Additional works \cite{hertz2023prompt, Chefer:2023} manipulate the cross-attention features combined with text, to affect specific image parts.
\citet{patashnik2023localizing} and \citet{dahary2024yourself} manipulate the self- and cross-attention layers to control the layout of the image and avoid semantic leakage between its different parts.
Others~\cite{geyer2024tokenflow,patashnik2024consolidating,hertz2024style} %
use combinations of $Q$, $K$, and $V$ for temporally and structurally consistent image and video editing.

Only a small number of works in the motion domain leverage the potential of attention editing. MoMo \cite{raab2024monkey} pioneers this direction by using patterns encoded in self-attention. MotionCLR \cite{chen2024motionclr} extends attention editing to cross-attention, and \citet{wu2025semantically} manipulates attention for motion stylization.

Our work, like MoMo, leverages attention features, but does so in a fundamentally different manner. 
We borrow ideas from Attend-And-Excite \cite{Chefer:2023}, which modulates cross-attention signals to strengthen underrepresented prompt tokens in the imaging domain.
Similarly, we manipulate cross-attention scores. However, to account for the substantial differences between the motion and image domains, we employ tailored, prompt-specific parameterization, defined by choices such as token and layer selection.

\section{Motion Generation}
\label{sec:Methodology}

In the next two sections, we describe \algoname, an unpaired, optimization based framework for improving text-to-motion alignment in composite multi-action scenarios. 
We show that cross-attention plays a key role in the ``vanishing semantics'' effect and use gradient-based guidance whose effectiveness depends on prompt-specific parameter settings. 
To this end, these parameters are determined with the help of 
a decision scheme, named \schemename, which outputs a set of parameters tailored to a given prompt. 

Our framework is built upon a diffusion-based \cite{ho2020denoising} generative backbone, implemented with a transformer-decoder \cite{vaswani2017attention}, that iteratively transforms a Gaussian noise $X_T$ into a clean motion $X_0$ conditioned on a text prompt $y$. 
Background on DDPM and $X_0$ prediction can be found in \appsecref{sec:mdm_ddpm_sup}{A.2}.

Our complete pipeline is illustrated in \cref{fig:pipeline}, where given an input text prompt, our framework uses \schemename to select the best parameters for that prompt (\cref{sec:param_select}), applies our customized generation procedure (\cref{sec:attn_guiding}), and outputs the resulting motion.

\subsection{Preliminaries}

\paragraph{Motion Representation.} Let $X \in \mathbb{R}^{N \times F}$ denote a motion sequence, where $N$ is the number of frames and $F$ represents the feature dimension per single frame. We follow the representation used in the HumanML3D dataset \cite{Guo:2022}, in which features from all joints (also known as pose) are concatenated into a single vector. Please refer to \appsecref{sec:represent_sup}{A.1} for more details. 

\paragraph{Cross-Attention Mechanism.}

Cross-attention plays a key role in our method, as it reflects the interaction between a motion latent and a textual prompt. 
Let $X_{t,\ell} \in \mathbb{R}^{N \times d_1}$ denote the motion feature tensor at diffusion step $t$ and transformer layer $\ell$, and let $E \in \mathbb{R}^{M \times d_2}$ denote the textual embedding of a prompt $y$ with $M$ tokens, where $d1,d2$ are latent dimensions. The cross-attention score $A$ is computed by
\begin{equation}
A = \text{Softmax}\left(\frac{Q K^T}{\sqrt{d}}\right), \quad Q = X_{t,\ell} W_Q^T, \quad K = E W_K^T,
\end{equation}
where $A \in \mathbb{R}^{N \times M}$ is the attention matrix for $N$ frames and $M$ tokens, $d$ is its dimension, and $W_Q,W_K$ are learned weights. 
The score $A_{:, m}$ reflects the influence of token $ m$ on the generated sequence.

To recover omitted actions (or action-details), we use an optimization strategy that explicitly increases the attention score of underrepresented tokens, as described next.

\subsection{Attention-Guided Generation} \label{sec:attn_guiding}

We define an \textit{underrepresented token} as a text token with low cross-attention and weak influence on the generated motion.
Next, we define an \textit{attention-alignment loss}, $\Loss_{atn}$, based on such tokens.

Given an underrepresented token index $m$, 
along with a specific diffusion step and transformer layer selected by our parameter-selection scheme (described later), we define:
\begin{equation} \label{eq:loss}
\Loss_{atn} = \frac{1}{N} \sum_{i=1}^{N} (1 - A_{i,m})^2,
\end{equation}

where $i$ is the frame number. This loss encourages the model to amplify attention to the underrepresented token along the temporal dimension. We update the motion latent $X$ via gradient descent:
\begin{equation}
X' = X - \eta \nabla_{X} \mathcal{L}_{atn},
\end{equation}
where $\eta$ is the refinement step size. 
This guidance steers the motion structure toward the textual semantics.
See also \cref{alg:attention_refine}.

\begin{figure}[t]
    \centering
    \includegraphics[width=\linewidth]{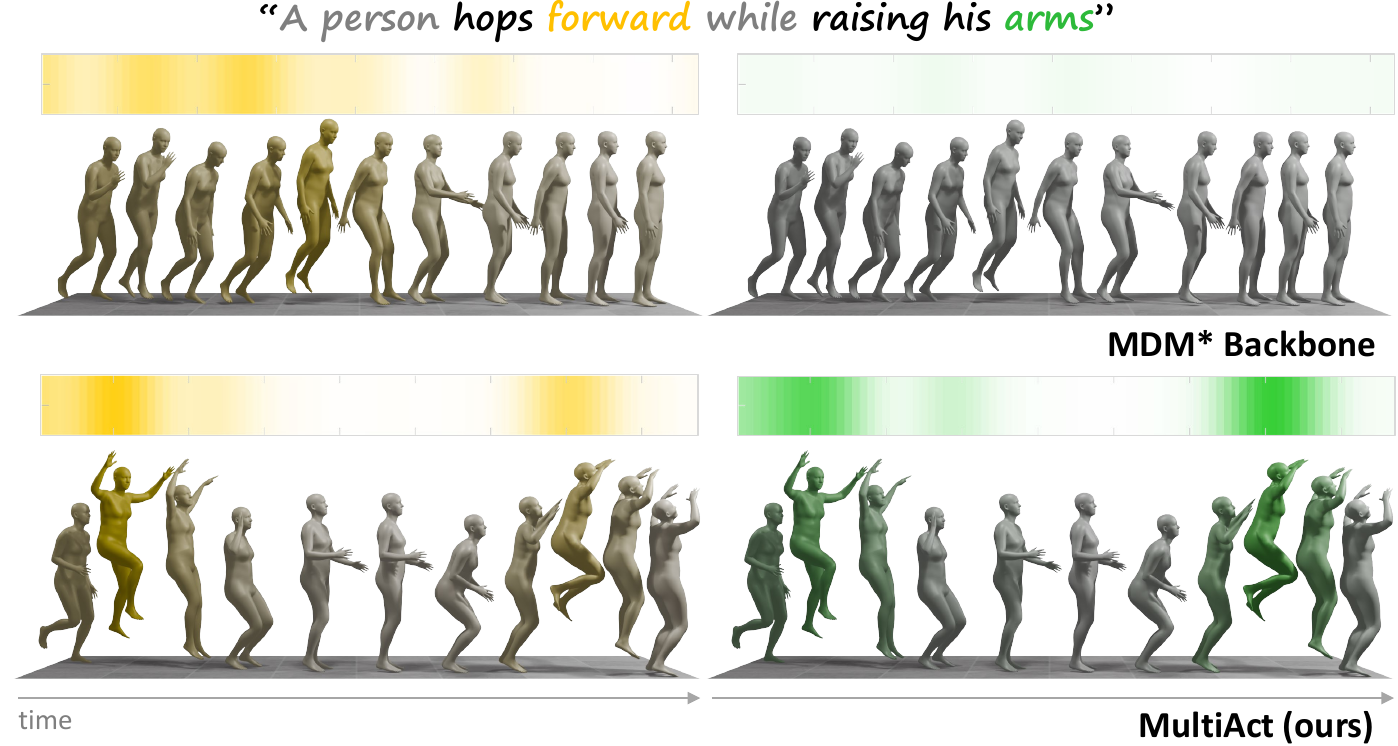}
    \caption{\textbf{Attention visualization.} The colored heatmaps illustrate attention scores for the words “forward” (yellow) and “arms” (green). Our backbone assigns low attention to arm-related tokens, resulting in motions in which the arms are not raised. In contrast, our method assigns high attention scores to both tokens, producing a synchronized motion that faithfully reflects the prompt. 
    }
    \label{fig:AttentionScore}
    \Description[Attention heatmaps before and after MultiAct guidance]{
    The figure visualizes attention over motion frames for selected prompt words. The backbone gives weak attention to the arm-related token and misses the arm motion, while MultiAct increases attention to both selected tokens and produces the intended composite motion.}
\end{figure}

\Cref{fig:AttentionScore} and our supplementary video visualize the effect of 
attention guidance on the attention score and the generated motion.

\section{Tailored Parameter Selection} \label{sec:param_select}

The efficacy of modulation in \cref{sec:attn_guiding} depends on the selection of appropriate parameters that ensure a meaningful alignment with the textual condition. Unfortunately, a configuration that succeeds for one prompt may fail for another. We demonstrate this dependency in our ablation study at \cref{sec:ablation}.

To achieve consistent high-fidelity results on a diverse range of prompts, we propose a decision scheme named \schemename. 
Our scheme gets a conditioning text prompt as input and 
outputs a set of parameters tailored to that prompt.
The selected parameters include the text token to amplify, the transformer layer used for guidance, and the number of diffusion steps over which the optimization is performed.
Note that we strengthen only one token, which we find sufficient.
This can be intuitively attributed to language embeddings, where a single token encodes other prompts via self-attention.

\begin{figure}[t]
    \centering
    \includegraphics[width=\linewidth]{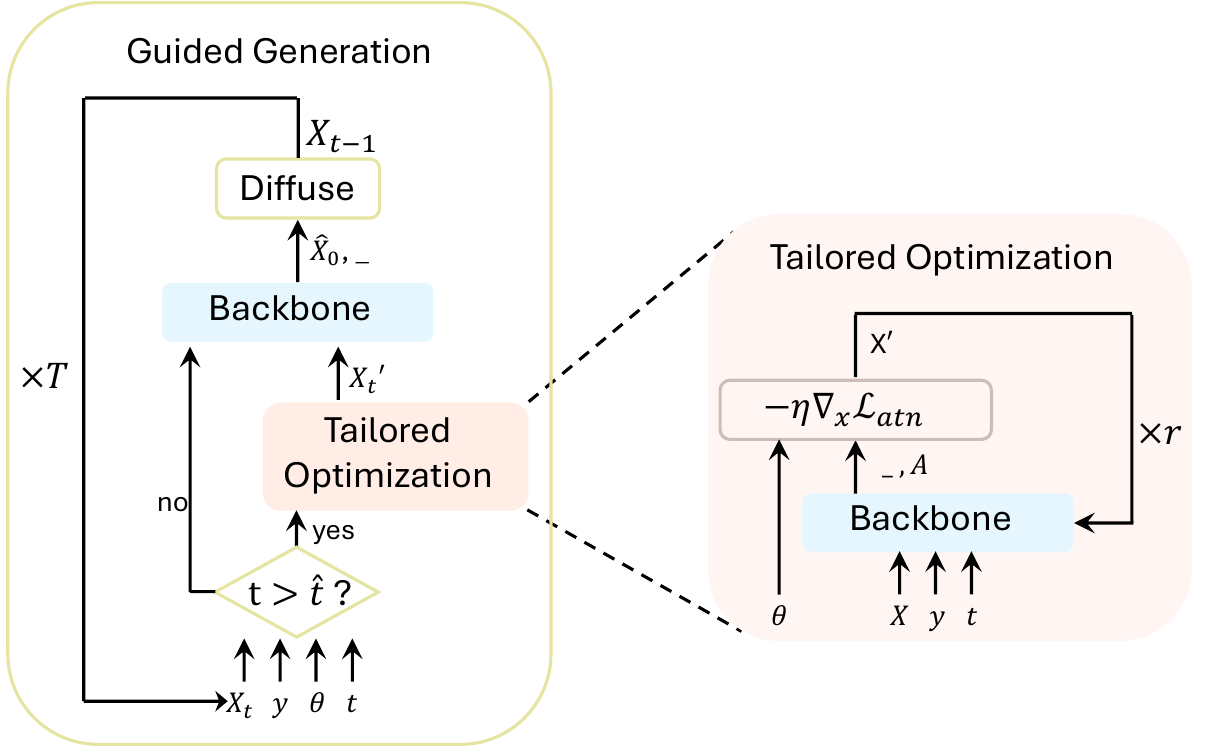}
    \caption{\textbf{Attention guided generation.} This figure visualizes \cref{alg:attention_refine,alg:guided_generation}. Left: Diffusion inference pipeline (\appsecref{sec:mdm_ddpm_sup}{A.2}) with tailored optimization integrated. Tailored optimization modifies the data tensors $X_t$ and is applied only 
    for $\hat{t}$ 
    diffusion steps, where $\hat{t}$ is tailored to the prompt $y$. Right: Tailored optimization, in which $X$ is adjusted to increase cross-attention for a selected token and layer, encapsulated in $\theta$ and also tailored to the prompt. The optimization is driven by the loss $\Loss_{atn}$ [\cref{eq:loss}], which is parameterized by the selected token and layer. $\eta$ and $r$ are the optimization step size and number of iterations, respectively.}

    \label{fig:guided_gen}
    \Description[Diagram of tailored attention-guided generation]{
    The figure illustrates where tailored optimization is inserted into the diffusion inference loop and how the optimization adjusts the current motion sample to increase attention for a selected token and transformer layer.}
\end{figure}

Once the most effective parameters have been identified, we run guided generation (\cref{sec:attn_guiding}), as shown in \cref{alg:attention_refine,alg:guided_generation,fig:guided_gen}.

In the following, we describe the components used by our decision scheme, namely the deviation measure, the parameter space, and the data. We then present the scheme in detail.

\paragraph{Notation.}
We use $motion(y,\theta)$ to denote the motion generated with attention guidance of input prompt $y$ and parameter combination $\theta$.
We use $err(y,\theta)$ to denote the deviation measure (described next) computed for $motion(y,\theta)$.

\subsection{Text Alignment Deviation Measure} \label{sec:deviation_measure}

\begin{figure}[t]
    \centering
    \includegraphics[width=\linewidth]{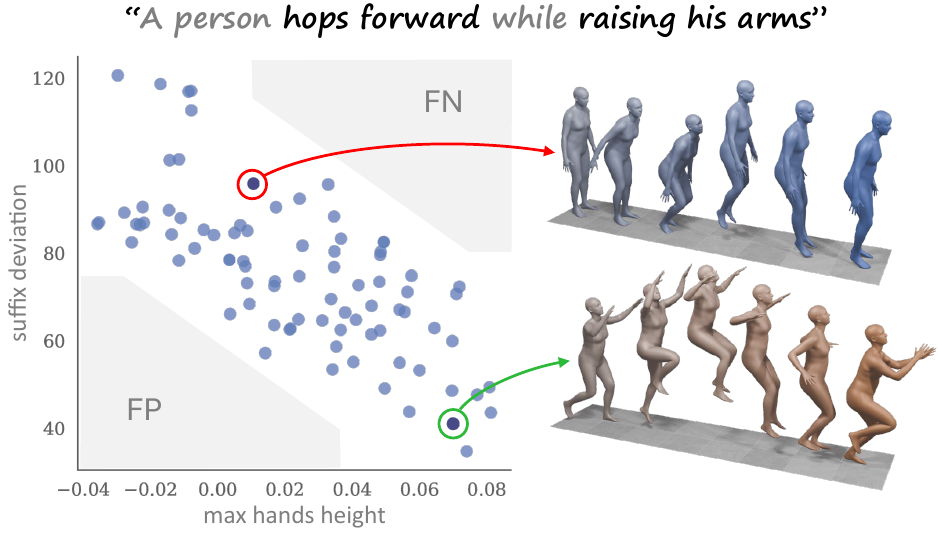}
        \caption{\textbf{Deviation measure credibility.} 
        Visual sanity check for the deviation scores, where each point corresponds to a parameter combination; the vertical axis shows the suffix deviation, and the horizontal axis depicts a prompt-specific geometric heuristic based on maximum hand-shoulder distance.        
        Low suffix deviation errors (\eg, green-circled) indicate correct generation of the suffix action (arm raising), which is also reflected by large hand-height values. In contrast, high errors (\eg, red-circled) correspond to failures. The absence of false positives (FP) and false negatives (FN) demonstrates the effectiveness of our measure in evaluating parameter combinations based on their degree of text–motion alignment.}
    \label{fig:MetricCredability}
    \Description[Scatter plot validating the deviation measure]{
    The scatter plot compares suffix deviation against a geometric hand-height heuristic for many parameter combinations. Low deviation points align with successful arm raising, high deviation points align with failures, and no false-positive or false-negative regions are marked.}
\end{figure}

Deviation values quantify the discrepancy between text and motion semantics, where smaller deviations correspond to stronger text–motion alignment, as illustrated in the sanity check in \cref{fig:MetricCredability}. We build our deviation estimator on the established multi-modal distance measure introduced in T2M \cite{Guo:2022}, which computes the distance between motion and text embeddings. Our goal is to minimize the deviation.

For convenience, we adopt the textual format \textit{``<prefix> while <suffix>''} (\eg \textit{a person is dancing while moving backwards}). 
However, \algoname is not restricted to this format and can operate on arbitrary syntactic structures. 
We evaluate prefix and suffix separately to weight all sentence parts equally.
The final error is computed as the L2 norm of the prefix and suffix deviations. Since our deviation is a combination of two measures, we call it \emph{dual multi-modal distance}.

\subsection{Parameter Space Construction} \label{sec:param_space}

\begin{algorithm}[t]
\caption{\textbf{Tailored attention-guided optimization.} }
\label{alg:attention_refine}
\begin{algorithmic}[1]
\STATE \textbf{Function} \textsc{TailoredAttnOptim}$(y, \theta, \phi, G, X, t)$
\STATE \parbox[t]{0.9\linewidth}{\textbf{Input:} Prompt $y$, predicted params. $\theta$, fixed params. $\phi$, data tensor $X$, current diffusion step $t$, where $\theta=\{m, \ell, \hat{t}\}$ denotes the selected token index, transformer layer, and last diffusion guidance step, and $\phi=\{\eta, r\}$ denotes the optimization step size and optimization steps, respectively}
\STATE \textbf{Output:} optimized tensor $X$
\IF{$t \le \hat{t}$}
    \RETURN $X$
\ENDIF
\FOR{$r$ iterations}
    \STATE $\_, A \gets G(X, y, t)$ \COMMENT{$A=[A_1,\dots,A_{L-1}]$, where $L$ is \#layers}
    \STATE $\mathcal{L}_{atn} \gets$ mean$[(1-A_\ell[:,m])^2]$
    \STATE $X \gets X - \eta \nabla_{X} \mathcal{L}_{atn}$ 
\ENDFOR
\RETURN $X$
\end{algorithmic}
\end{algorithm}

\begin{algorithm}[t]
\caption{\textbf{Tailored attention-guided generation.} Diffusion process with tailored attention guidance. We omit details such as prompt embeddings and layers surrounding the transformer in $G$.}
\label{alg:guided_generation}
\begin{algorithmic}[1]
\STATE \textbf{Function} \textsc{GuidedGeneration}$(y, \theta, \phi, G, T)$
\STATE \parbox[t]{0.9\linewidth}{\textbf{Input:} Prompt $y$, predicted params. $\theta$, fixed params. $\phi$, pretrained diffusion model $G$, and diffusion steps $T$}
\STATE \textbf{Output:} Generated motion $x_0$
\STATE $X \gets \mathcal{N}(0, 1)$ 
\FOR{$t \gets T, T-1, \dots, 1$}
    \STATE $X' \gets$ \textsc{TailoredAttnOptim}$(y, \theta, \phi, G, X, t)$ \COMMENT{Alg.~\ref{alg:attention_refine}}
    \STATE $\hat{X}_0, \_ \gets G(X', y, t)$ \COMMENT{each step $t$ predicts a clean $\hat{X}_0$}
    \STATE $X \gets$ \textsc{Diffuse}$(\hat{X}_0, t)$ \COMMENT{noise $\hat{X}_0$ to get $x_{t-1}$}
\ENDFOR
\RETURN $X$
\end{algorithmic}
\end{algorithm}

We construct a parameter set that encompasses combinations of tokens, layers, and diffusion ranges. This parameter space is large, making the selection of an optimal parameter combination for a given text prompt challenging. We therefore address parameter selection in two stages.
First, we narrow the candidate parameter space using clustering-based pruning, motivated by the clear visual separation observed in the deviation plots, which produces a reduced parameter set $\Theta$.
Second, we employ a decision scheme described in \cref{sec:param_gate}, which applies lightweight non–deep-learning methods to identify the optimal set of guidance parameters.

A text token index is denoted by $m \in [0,\dots,M\!-\!1]$, where $M$ is the number of tokens in a given prompt, and 
a transformer layer is denoted by $\ell \in [0,\dots,L\!-\!1]$, where $L$ is the number of layers. 
The number of diffusion steps over which guidance is applied is denoted by $\hat{t}$, corresponding to the first $\hat{t}$ steps starting from step $T$, where $T$ is the total number of diffusion steps.
Hence, the preliminary parameter set can be written as: 
\begin{equation}
    \Theta_\text{{pre}}\!=\!\{m,\ell,\hat{t} | 
    m \mathrel{\!\in\!} [0 \dots M\!-\!1], 
    \ell\mathrel{\!\in\!} [0 \dots L\!-\!1], 
    \hat{t} \mathrel{\!\in\!} [T\!-\!1 \dots 0]\}. %
\end{equation}
To narrow this space, we examine deviation values of motions generated for $\theta \in \Theta$ across several text prompts, revealing a clear separation in which only a small subset of $\Theta$ yields low deviation.

\begin{figure}[t]
    \centering
    \includegraphics[width=1\linewidth]{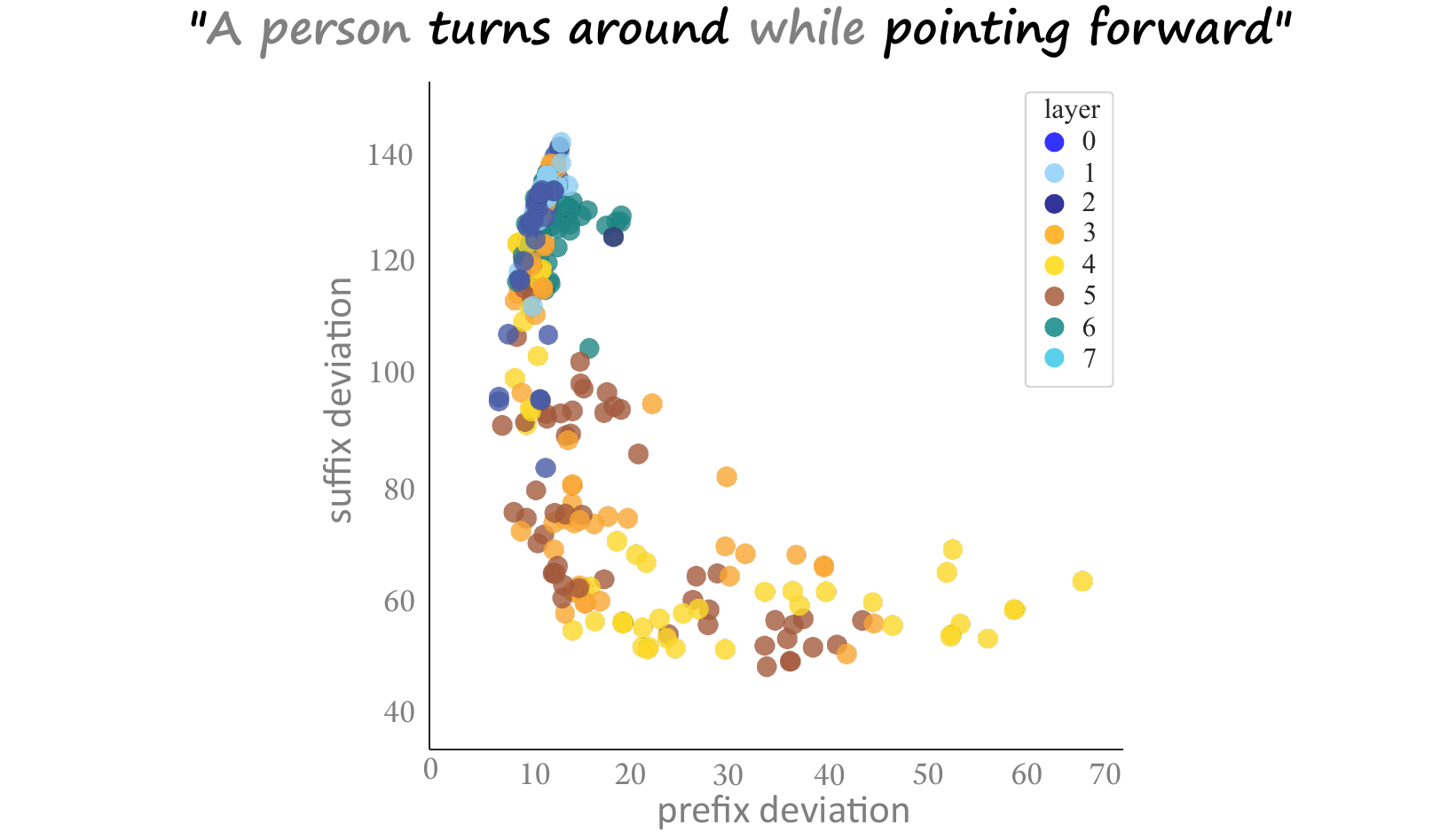}
        \caption{
        \textbf{Deviation space visualization}. Each point corresponds to a single parameter combination and shows the prefix and suffix deviation values for the prompt shown above. 
        Distance from the origin indicates the dual multi-modal distance. 
        Points are color-coded by transformer layer $\ell$, revealing a visual separation in which parameter combinations closest to the origin are dominated by layers $3$ to $5$.
        }
    \label{fig:ExhaustiveSearch}
    \Description[Deviation space plot colored by transformer layer]{
    The scatter plot places each parameter combination according to prefix and suffix deviation. Points closer to the origin have lower dual deviation, and the color coding shows that the strongest combinations cluster around specific transformer layers.}
\end{figure}

\Cref{fig:ExhaustiveSearch} illustrates the deviation space in general and the visual separation between layers in particular. 
Each point represents the deviation value for the prefix and suffix of a single motion instance from the set ${motion(y,\theta)}_{\theta \in \Theta}$, generated for a fixed prompt $y$ (“turns around while pointing forward”) across all combinations of parameters.
The distance of a point from the origin corresponds to $err(y,\theta)$ (the dual multi-modal distance).
The deviation space forms a curved manifold that reflects a clear trade-off: improving alignment for one part of the sentence typically corresponds to reduced alignment for the other. The region of interest, therefore, lies near the origin, where both sentence parts are well aligned with the prompt.
While the figure shows the deviation space for a single text prompt, a similar structure consistently appears across different prompts.

The instances in \cref{fig:ExhaustiveSearch} are color-coded according to the layer $\ell$ in the parameter combination $(m,\ell,\hat{t})$.
We observe a clear visual separation, with the best text-aligned parameter combinations near the origin dominated by layers $3$ to $5$.

To further support this observation, we use a small set of prompts, $\Ycal_{\text{pre}}$, and exhaustively generate attention-guided motions 
$\Ical_{\text{pre}}$, together with their deviation scores $\Ecal_{\text{pre}}$:

\begin{equation}
\left.
\begin{aligned}
    \Ical_{\text{pre}} &= \{\, motion(y,\theta) \,\}, \\
    \Ecal_{\text{pre}} &= \{\, err(y,\theta) \,\}
    \end{aligned}
\;\right|\;
y \in \Ycal_{\text{pre}},\ \theta \in \Theta_{\text{pre}}.
\end{equation}
Within $\Ecal_{\text{pre}}$, we cluster samples by their distance to the origin using k-means (k=2) and assign each layer to the “near” or “far” group only when the fraction of its samples in a cluster exceeds a confidence threshold; layers that do not meet this margin are labeled ambiguous. We repeat the procedure for tokens and steps too.
This process narrows the set of text token indices to $\Mcal$, candidate layers to $\Scal$, and diffusion step ranges to $\Tcal$, all detailed in \cref{sec:imp_details}.
The narrowed sets are no longer separable via clustering and therefore require a more sophisticated scheme, which is provided by \schemename.
Altogether, our parameter space is narrowed down to 
\begin{equation}
    \Theta=\{ m,\ell,\hat{t} | m \in \Mcal, \ell \in \Scal, \hat{t} \in \Tcal \}.
\end{equation}

\subsection{Data Preparation} \label{sec:data_prep}

We construct a set of prompts, $\Ycal$, of plausible co-occurring actions (\eg, “running while squatting” is not plausible). See \cref{sec:imp_details} for details.
For each prompt and each parameter combination, we generate a motion, resulting in $\sim$30K motions.
Finally, for each generated motion, we compute a text-alignment deviation error. Altogether: %
\begin{equation}
\left.
\begin{aligned}
    \Ical &= \{\, motion(y,\theta) \,\}, \\
    \Ecal &= \{\, err(y,\theta) \,\}
    \end{aligned}
\;\right|\;
y \in \Ycal,\ \theta \in \Theta.
\end{equation}

The values in $\Ecal$ then serve as a guide to our decision scheme.
We gather all the above into a dataset, which we denote 
\begin{equation}
\Dcal=\{\Ycal,\Theta,\Ical,\Ecal\}.
\end{equation}

\subsection{Decision Scheme} \label{sec:param_gate}

Let $\widebar{\theta}(y)$ denote the parameter combination $\theta=(m,\ell,\hat{t})$ that yields the lowest deviation for a text prompt $y$: 
\begin{equation}
    \widebar{\theta}(y)=\argmin_{\theta \in \Theta}err(y,\theta).
\end{equation}

We define the parameters and error extracted from $\bar{\theta}(y)$ as 
\begin{equation}
\begin{aligned}
\widebar{m}(y) &= \bar{\theta}(y)[m], \\
\widebar{\ell}(y) &= \bar{\theta}(y)[\ell], \\
\widebar{t}(y) &= \bar{\theta}(y)[\hat{t}], \\
\widebar{err}(y) &= err(y,\bar{\theta}(y)).
\end{aligned}
\end{equation}

Given an unseen text prompt $y_{new} \notin \Ycal$, we use our decision scheme, \schemename, to predict $\widebar{\theta}(y_{new})$.
Recall that $\Ycal$ is constructed under restrictive criteria, resulting in 100–200 prompts. Given the size of $\Ycal$, we employ lightweight, non-deep learning-based methods.

We analyze the nature of each parameter type and, based on this analysis, define a prediction strategy for each.
The decision procedures defined below are evaluated in a cross-validation manner on the dataset $\Dcal$. In \cref{sec:quant_res} we show that they are suitable choices.

\paragraph{Selecting $\ell$.}
To predict the layer $\ell$, we analyze the set $\{\widebar{\ell}(y)\}_{y\in\Ycal}$ and 
observe that the optimal guidance layer is shared across semantically similar prompts.
Motivated by this observation, 
we employ a nearest-neighbor approach, in which unseen prompts adopt the layer assignment of the most similar prompt, identified as

\begin{equation}
    \ell=\bar{\ell}(y_{nn}),  \quad\quad
    y_{nn}=\text{argmin}_{y\in\mathcal{Y}}\text{diff}(y_{new},y), 
\end{equation}

where $\text{diff}(\cdot,\cdot)$ is the L2 distance between CLS token embeddings.

\begin{figure*}[!t]
    \centering
    \includegraphics[width=\linewidth]{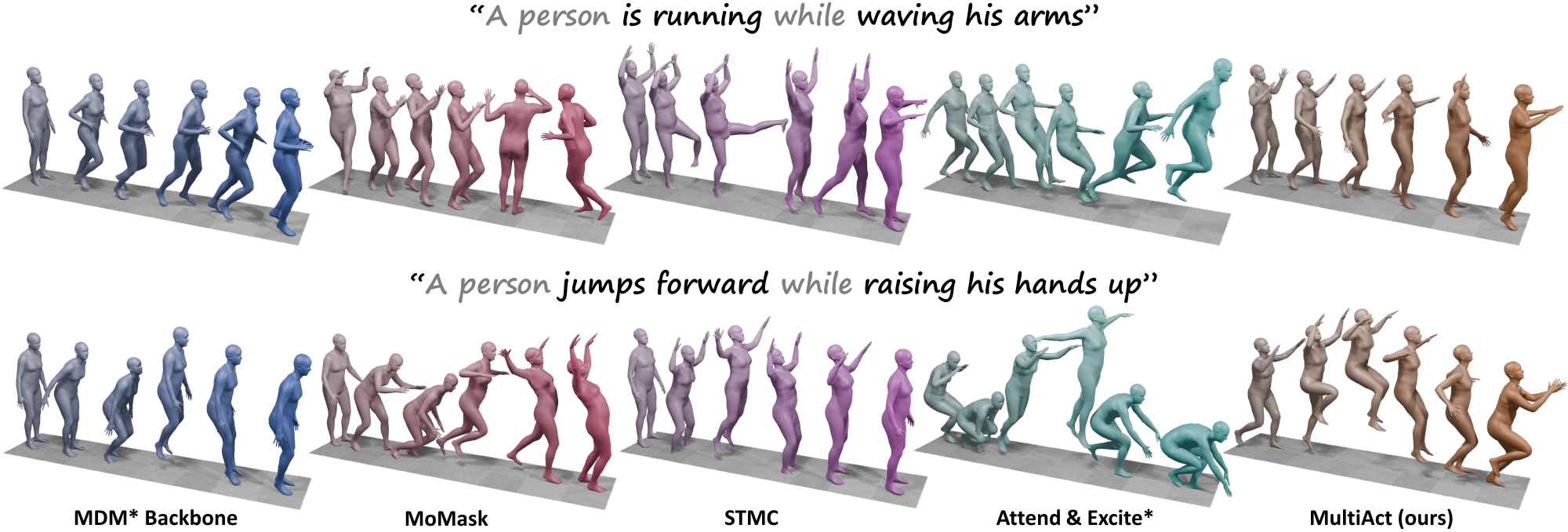}
    \caption{\textbf{Qualitative results.} Our method consistently maintains high motion quality and strong text alignment.
    In contrast, other baselines either attend to a single dominant verb, exhibit occasional floor penetration or motion artifacts, or produce unnatural results. }
    \label{fig:qualitative}
    \Description[Qualitative comparison across several text-to-motion methods]{
    The figure compares motions generated from composite prompts by MultiAct and several baselines. MultiAct preserves both requested action components, while the baselines often miss one action, show artifacts, or produce less natural motion.}
\end{figure*}

\paragraph{Selecting $\hat{t}$.}
To predict $\hat{t}$, the number of diffusion steps over which optimization is performed, we test the hypothesis that larger deviations require guidance over more diffusion steps. 
To do so, we examine the set of tuples $\{\widebar{t}(y), err(y)\}_{y\in\Ycal}$. Note that $err(y)$ is the backbone error, not the error $\widebar{err}(y)$ of the optimal $\theta$.
We identify thresholds,
$\{\kappa_i\}_{i\in[0\dots\abs{\Tcal}\!-\!2]}$,
by empirical risk minimization over 1D ordered classifiers using cumulative label counts.

We generate $motion(y_{new})$ using the backbone (no $\theta$ required when using the backbone), compute the deviation error $err(y_{new})$, and predict $\hat{t}$ using the previously detected thresholds:
\begin{equation} 
    \hat{t} =
    \begin{cases}
        t_0, & \text{if } err(y_{new}) \le \kappa_0, \\
        t_1, & \text{if } \kappa_0 < err(y_{new}) \le \kappa_1, \\
        \smash{\vdots} & \\[-5pt]
        t_{\abs{\Tcal}-1}, & \text{if } \kappa_{\abs{\Tcal}-2} < err(y_{new}), %
    \end{cases}
\end{equation}
where $t_0>t_1> \dots > t_{\abs{\Tcal}-1}$, and $\kappa_0<\kappa_1< \dots < \kappa_{\abs{\Tcal}-2}$.

\paragraph{Selecting $m$.}
Finally, to predict $m$, the token index in the input prompt, we propose test-time scaling. We use an LLM to identify all the verbs, action-details and adjectives in $y_{new}$ (\ie, ignore non-informative tokens like ``person'' and ``his'').
This process typically yields 3-4 candidate tokens.
For each candidate, we generate a motion and evaluate its deviation measure. We then select the token corresponding to the minimal deviation.
All qualitative results in this work use a token $m$ selected by this procedure.
Notably, we also suggest a faster but less accurate method where we simply sample a token out of the narrowed set $\Mcal$. Our quantitative results (\cref{sec:quant_res}) show that both methods outperform existing baselines.

\section{Experiments} \label{sec:exp}

\subsection{Implementation Details} \label{sec:imp_details}

For our optimization procedure, we set the size of the optimization step, $\eta$, to $0.02$, and the number or repetition of optimization for each denoising step is $r=$ $18$.

Our narrowed down parameter spaces, $\Mcal$, $\Scal$, and $\Tcal$ are defined as follows. Recall that they are selected from the full parameter spaces using a simple clustering method.
The candidate layers for the optimization process are layers $3$ to $5$ (out of $8$ layers in total); 
$\hat{t}$ should be chosen such that modulation would be applied only during the early, structure-forming steps of the reverse diffusion process. In practice, we narrow down $\hat{t}$ to $46$ to $48$ (out of $50$ steps), i.e., optimization is applied for $1$ to $3$ diffusion steps.

Our cluster analysis shows that the best performing tokens $\Mcal$ are those classified as action-details (e.g., adverbs, directions, or manners) rather than verbs. 
This observation aligns with the intuition that in self-attention, information is shared across tokens during encoding. We hypothesize that action-detail tokens tend to encode both the details and the underlying action, whereas verb tokens mainly represent coarse action semantics. Consequently, action-detail tokens are more informative and better guide the prediction.
Altogether, 
$\Mcal= [\text{action-details}], \Scal= [3,4,5], \Tcal=[48,47,46]$.

When measuring text alignment deviation errors, 
we determine each deviation value to be the average of deviation values for eight motions, $motion(y,\theta)$, all created from the same prompt $y$ and the same parameter combination $\theta$, but with different random seeds.

\paragraph{Prompt set construction.}
Our prompt sets, $\mathcal{Y}_{pre}$ and $\mathcal{Y}$, contain hard yet clear examples: challenging for most baselines but textually unambiguous (akin to GenEval \cite{ghosh2023geneval} for images). Prompts must (a) be clear and definite, (b) have components that most baselines can generate individually, (c) combine components that most baselines struggle to generate jointly, (d) be human-doable, and (e) describe simultaneous (not sequential) actions.

The preliminary set $\mathcal{Y}_{pre}$ contains 15 manually designed prompts with diverse actions and body parts, typically combining locomotion with instantaneous actions (e.g., punch while walking). It is used for initial analysis, focusing on motions with strong temporal signals that are easy to evaluate visually and via attention (e.g., arm lifting in \cref{fig:MetricCredability,fig:AttentionScore}).

The final prompt set $\mathcal{Y}$ contains 140 prompts, generated with ChatGPT-5.2 using examples from (a) HumanML3D prompts of the form ``<prefix> while <suffix>”, and (b)  $\mathcal{Y}_{pre}$. The (shortened) instruction is:
``Create a sentence in the spirit of the examples, of the form ‘a person <prefix> while <suffix>’, where both parts contain actions and action details and can realistically occur simultaneously.”

Both sets are refined through iterative filtering: prompts are generated, motions synthesized by the backbone, easy cases removed, and harder variants added. For example, while the backbone successfully handles an action co-occurring with forward locomotion, it struggles when the same action co-occurs with backward locomotion; therefore, we retain the latter.

\paragraph{Backbone.} 
Our denoising backbone is a modified version of MDM \cite{Tevet:2023:MDM}, retrained on the HumanML3D dataset \cite{Guo:2022} and denoted MDM*. We use MDM’s transformer-decoder variant with text conditioning on BERT embeddings \cite{devlin-etal-2019-bert}, where motion frames attend to each embedded text token.

Adapting MultiAct to a new backbone requires a one-time construction of the motion set $\Ical\!=\!{motion(y,\theta)}_{y \in \Ycal,\theta \in \Theta}$ (approximately $4\mathrm{K}$ generations), which takes about two hours with 4 GPUs and a batch size of 8.

\paragraph{Inference time.}
MultiAct incurs a $\sim\times 2$ overhead relative to backbone inference time due to the optimization loop. Applying test-time scaling introduces an additional $\sim\times 3$ factor (for 2–3 tokens), resulting in an overall of $\sim\times 6$ the backbone runtime. For comparison, STMC runs separately per body part, incurring a $\sim\times 5$ overhead.

\begin{figure*}[t]
    \centering
    \includegraphics[width=\linewidth]{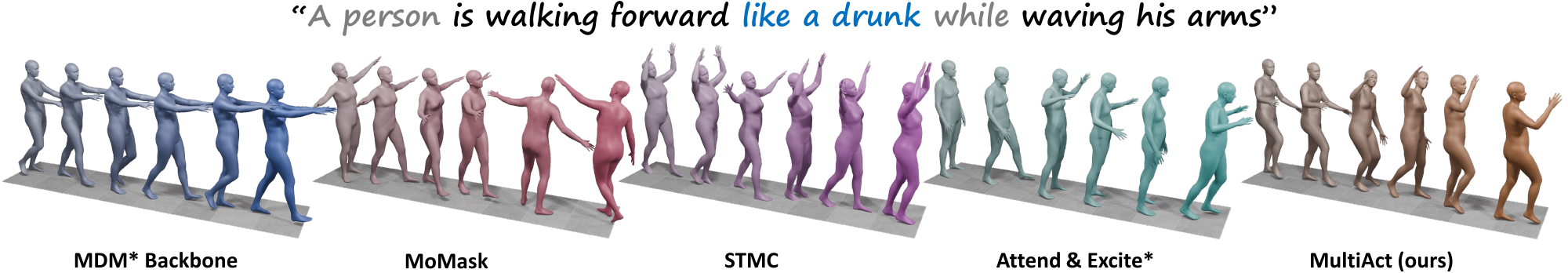}
    \caption{\textbf{Motion stylization}. When prompted with motion stylization, existing methods fail either to generate both actions simultaneously or to incorporate the requested style. In contrast, our method supports motion stylization in parallel with multiple simultaneous actions, successfully integrating both into the generated motion. Note the characteristic imbalance in our result, reflecting the specified drunken style.}
    \label{fig:MotionStyle}
    \Description[Comparison of styled composite motion generation]{
    The figure compares generated motions for a prompt that combines simultaneous actions with a requested motion style. MultiAct follows both the actions and the style, while the baselines either omit actions or fail to express the style.}
\end{figure*}

\begin{figure*}[t]
    \centering
    \includegraphics[width=0.8\linewidth]{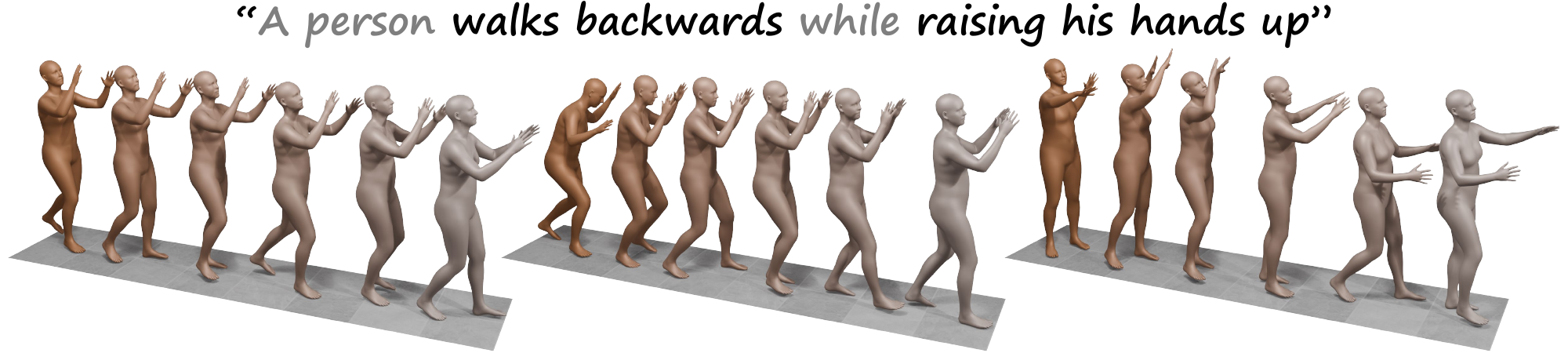}
    \caption{\textbf{Motion diversity}. When sampled multiple times with the same prompt and parameter combinations, \algoname generates diverse, high-quality motions that consistently satisfy both simultaneous actions.}
    \label{fig:MotionDiversity}
    \Description[Diverse samples from the same composite motion prompt]{
    The figure shows multiple generated motions sampled from the same prompt and guidance parameters. The poses differ across samples while consistently satisfying the simultaneous actions in the text.}
\end{figure*}

\begin{table*}[t]
\centering
\caption{
\textbf{Comparison with baselines}. \algoname consistently outperforms the baselines across all categories on composite prompts.
STMC, designed to accommodate co-occurring actions, ranks second in the user study.
User study results reported here reflect \cref{fig:user_study}, where our score is the average over comparisons with all baselines. \textbf{Bold} and \underline{underline} denote best and second best, respectively. $(\cdot)^*$ indicates the method was adapted to align with our task.}
\label{tab:quant_compare}
\Description[Baseline comparison table for motion generation]{
The table compares MDM, MoMask, STMC, Attend and Excite, and two MultiAct variants using R precision, dual multi-modal distance, and user-study preference scores. MultiAct with test-time scaling has the best dual multi-modal distance and user-study scores.}
{\small
\begin{tabular}{lccccccc}
\toprule
\multirow{2}{*}{Method}  &   \multicolumn{3}{c}{R Precision $\uparrow$} &  \multirow{2}{*}{Dual MM Dist $\downarrow$}  & \multicolumn{3}{c}{User Study $\uparrow$} \\ 
\cmidrule(lr){2-4}  \cmidrule(lr){6-8}
                                            & Top1         & Top2         & Top3      &            & Text Align.   & Quality  & Overall Pref.  \\ \midrule
MDM* \citeyearpar{Tevet:2023:MDM}           & 0.08         & \underline{0.22}         & \textbf{0.32} & 110.64     & 2.49          & 13.53    & 8.74           \\
MoMask \citeyearpar{guo2024momask}          & 0.03         & 0.05         & 0.08      & 134.34     & 13.71         & 13.22    & 14.56          \\
STMC \citeyearpar{Petrovich:2024}           & 0.07         & 0.17         & 0.17      & 104.60     & \underline{22.50}         & \underline{20.78}    & \underline{23.47}          \\
Attend \& Excite* \citeyearpar{Chefer:2023} & 0.03         & 0.05         & 0.08      & 116.44     & 13.54         & 17.71    & 20.00          \\
MultiAct (Ours)                             &              &              &           &            &               &          &                \\
\quad Action-detail                         & \textbf{0.14}         & 0.19         & \underline{0.27}         & \underline{96.07} & N/A            & N/A        &       N/A         \\
\quad Test-time Scaling                     & \underline{0.11}         & \textbf{0.30}         & \textbf{0.32}         & \textbf{85.16}    & \textbf{86.94}         & \textbf{83.69}    & \textbf{83.30}          \\ \bottomrule
\end{tabular}
} %
\end{table*}

\subsection{Quantitative Results} \label{sec:quant_res}

\cref{tab:quant_compare} compares \algoname with state-of-the-art baselines.

\paragraph{Baselines.} 
We compare our method against four baselines. Two of these are text-to-motion models that are not explicitly designed to handle simultaneous actions, while the remaining two incorporate mechanisms specifically tailored for simultaneous action modeling.
The first group includes our backbone model, MDM*, and MoMask \cite{guo2024momask}. We include these baselines to demonstrate that the limitations addressed by our approach are common across a variety of text-to-motion models, both diffusion-based (MDM*) and VQ-VAE–based (MoMask).
The latter group includes STMC \cite{Petrovich:2024} and a variation of Attend-and-Excite \cite{Chefer:2023}, adapted to the motion domain.

MoMask \cite{guo2024momask} represents a model class with an architecture that differs significantly from our backbone, MDM, as it uses a hierarchical vector-quantized token representation with masked and residual transformers. While both MDM and MoMask are considered state-of-the-art, their expressivity is limited, and our experiments show that they struggle with composite prompts.

STMC \cite{Petrovich:2024} handles both simultaneous and sequential motion generation, in which users are required to specify overlapping multi-track prompt intervals, based on body-part involvement. Since actions are assigned to body parts, the method struggles with simultaneous combinations of full-body and part-specific actions.
STMC cannot handle adjectival modifiers (e.g., “a drunk person walks”), which constitutes a limitation of the method. To ensure a fair comparison, we reformulate our prompts to match the STMC input format (e.g., “a person walks like a drunk”). Even under this constraint, our method outperforms theirs.

Attend-and-Excite \cite{Chefer:2023} is the method most closely related to our work, as it aims to strengthen specific textual tokens through optimization. With minimal modifications, we adapt it to the motion domain to enable comparison, and name the adapted version Attend-and-Excite*. There are several key differences between their method and ours. 
First, Attend-and-Excite optimizes attention across all layers and 25 (out of 50) diffusion steps, whereas we restrict optimization to one selected layer and step. Second, while it maximizes attention towards a predefined target score, we instead perform a fixed number of optimization steps and optimize the mean attention score. 
Finally, Attend-and-Excite requires users to specify target tokens, whereas in the motion domain, users cannot reliably identify which tokens have the greatest effect; consequently, we introduce an automatic token-selection strategy. To ensure a fair comparison, we select the most appropriate tokens as their input. 
Our comparisons show that while their design choices are well-suited for images, they are not optimal for the motion domain.

\begin{figure}[t]
    \centering
    \includegraphics[width=.85\linewidth]{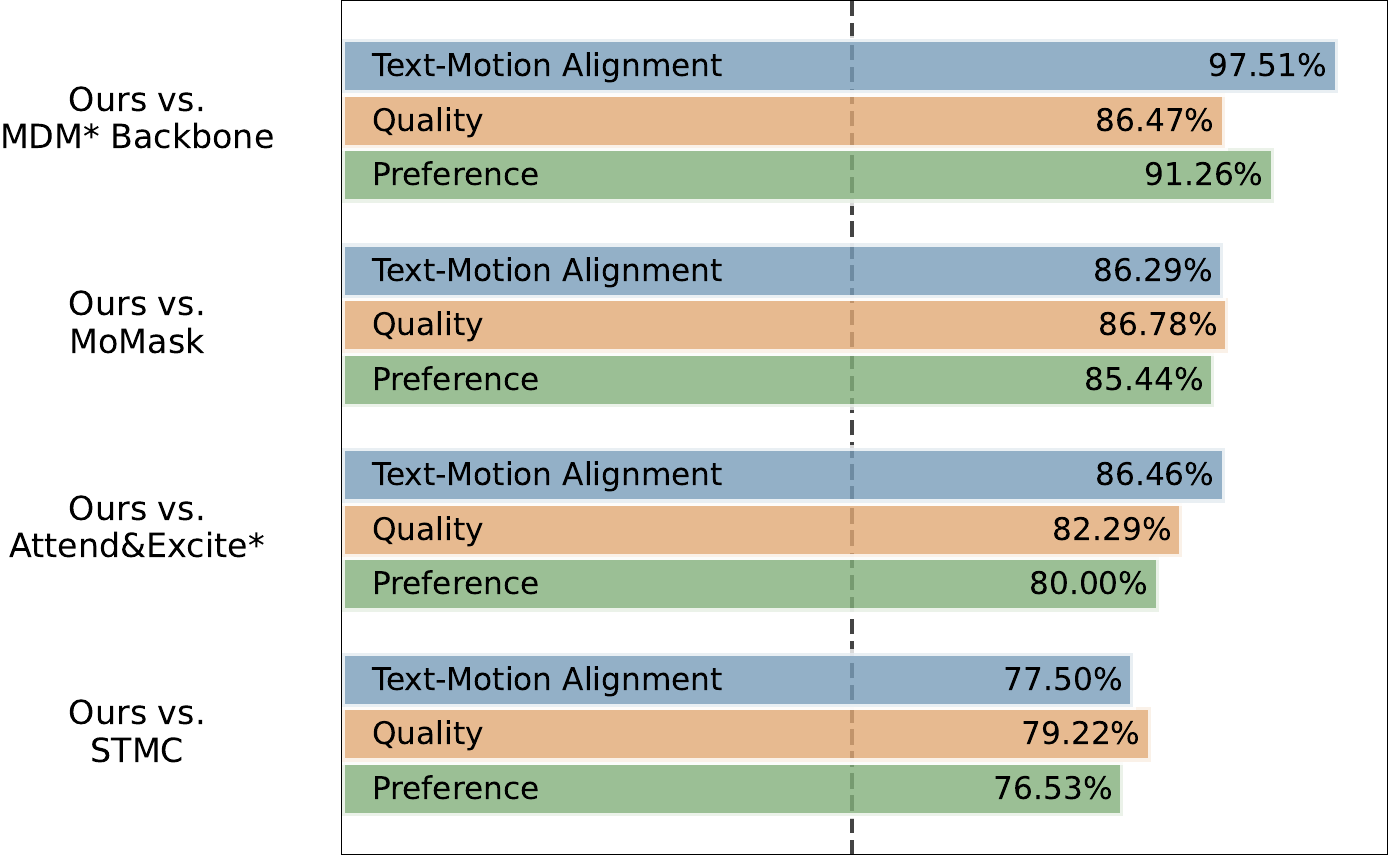}
    \caption{\textbf{User Study.} Users vote that our model performs better than state-of-the-art models. The percentage indicates the proportion of votes in favor of our method, and the dashed line denotes the 50\% mark. 
    }
    \label{fig:user_study}
    \Description[User study preference percentages comparing MultiAct to baselines]{
    The bar chart reports the percentage of user votes favoring MultiAct over each baseline, with a dashed line marking equal preference at fifty percent.}
\end{figure}

\paragraph{User Study.} 
We use the term ``user study'' for clarity, although ``perceptual study'' is more precise in this case.
We compare our framework with the baselines using three criteria: quality, text alignment, and overall preference. The user study comprises 28 comparisons in total, with 7 comparisons per baseline, each evaluated by 16-18 distinct users. A screenshot of our study can be found in \appsecref{sec:user_study_screenshot}{B.1}. \Cref{tab:quant_compare,fig:user_study} show that \algoname is preferred over the models compared across all evaluation criteria.
The prompts used in the user study are a small subset of $\Ycal$, chosen to be simple and easy to evaluate visually, allowing clear judgment of success.

\paragraph{Metrics.}
We evaluate both text alignment and motion quality.
For text alignment, we use R Precision, Dual MultiModal Distance \cite{Guo:2022} (where ``dual'' denotes the L2 norm of the prefix and suffix deviations), and the text-alignment score from our user study.

To assess motion quality, we use the quality and user preference scores from our user study.
Note that our dataset is unpaired; therefore, metrics such as FID and diversity are not applicable. 
Diversity, however, is demonstrated in our qualitative results.

\paragraph{HumanML3D discussion and results.}

At first glance, the subset of the HumanML3D test set following the “<prefix> while <suffix>” format could serve as a suitable test set.
However, this subset does not align with the objectives of our prompt set, which is designed to include examples that are both hard and clear (see \cref{sec:imp_details}).
The requirement for \textit{hard} prompts is not met, as approximately 40\% of the prompts are relatively easy (e.g., “raise hands while walking forward”) and can be generated by most baselines. As a result, they do not allow us to distinguish between stronger and weaker baselines. (40\% denotes prompts with backbone Dual MM-Dist below 50).
The requirement for \textit{clear} prompts is also not satisfied, as some prompts are linguistically complex and admit multiple interpretations (e.g., containing more than 50 words and multiple overlapping actions), leading to unreliable metric scores.
Nevertheless, we report results on this subset (539 prompts) in \Cref{tab:humanml_subset} and observe that MultiAct outperforms the backbone in all categories except one, in which it achieves comparable performance.

\begin{table}[t]
\centering
\caption{\textbf{Evaluation on HumanML3D ``<prefix> while <suffix>'' subset.} 
MultiAct outperforms its backbone, despite the test set being suboptimal for our purposes.}
\label{tab:humanml_subset}
\Description[HumanML3D subset evaluation comparing MDM and MultiAct]{
The table compares MDM and MultiAct on R precision, dual multi-modal distance, FID, and diversity for the HumanML3D prefix-while-suffix subset. MultiAct improves R precision, dual multi-modal distance, and FID, while MDM has slightly higher diversity.}

\small
\resizebox{\columnwidth}{!}{
\begin{tabular}{lcccccc}
\toprule
\multirow{2}{*}{Method}  
& \multicolumn{3}{c}{R Precision $\uparrow$} 
& \multirow{2}{*}{\begin{tabular}{c}Dual MM\\Dist $\downarrow$\end{tabular}} 
& \multirow{2}{*}{FID $\downarrow$}
& \multirow{2}{*}{Diversity $\rightarrow$} \\
\cmidrule(lr){2-4}  
& Top1 & Top2 & Top3 \\
\midrule
MDM* \citeyearpar{Tevet:2023:MDM} 
& 0.14 & 0.24 & 0.31 & 59.65 & 3.39 & \textbf{8.06} \\
MultiAct 
& \textbf{0.19} & \textbf{0.30} & \textbf{0.39} & \textbf{58.73} & \textbf{2.77} & 7.92 \\
\bottomrule
\end{tabular}
} %

\end{table}

\subsection{Qualitative Results} \label{sec:qual_res}

Our supplementary video best illustrates the quaity of our results, also shown in the following figures.
\Cref{fig:qualitative,fig:MotionStyle} compare our work with baselines. As shown in \Cref{fig:qualitative}, baselines struggle with simultaneous actions (MDM), produce unnatural motions or do not fully follow the text (MoMask), or exhibit floor penetration and minor artifacts (STMC). 
Attend-and-Excite proves ineffective for motion guidance, exhibiting corrupted motion that arises from the lack of prompt-tailored parameters, and applying attention guidance across all layers and over many diffusion steps.

\Cref{fig:MotionStyle} shows that, for composite prompts involving stylization, our method is the only one that successfully follows both the specified actions and the style. In contrast, STMC, which is designed to map actions to body parts, fails when combining full-body styles (“drunk”) with part-specific actions (“arms”).

Lastly, \cref{fig:MotionDiversity} shows that \algoname generates diverse motions using a selected parameter configuration with different seeds.

\begin{figure}[!t]
    \centering
    \includegraphics[width=\columnwidth]{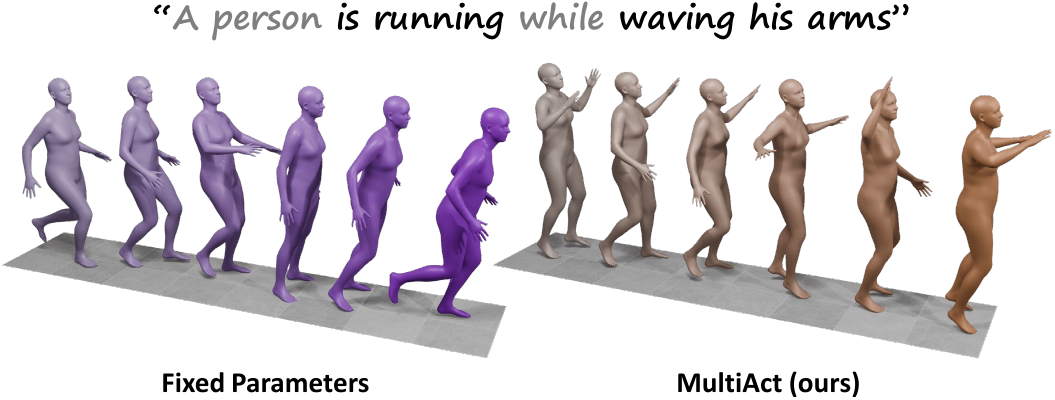}
    \caption{\textbf{Ablation.} 
    Left: Motion generated using a fixed parameter set selected according to the best-performing combination in $\Theta$. 
    Right: Motion generated for the same prompt with parameters selected by \schemename.}
    \label{fig:Ablation}
    \Description[Ablation comparison between fixed parameters and ParamGate]{
    The figure shows two motions for the same prompt. The fixed-parameter result fails to match the prompt as well as the result generated with prompt-tailored parameters selected by ParamGate.}
\end{figure}

\begin{table}[t]
\centering
\caption{\textbf{Ablation.} Using a fixed parameter set yields poor results, while progressively introducing prompt-tailored selection of layers, steps, and tokens improves alignment, with test-time scaling performing best.}
\label{tab:ablation}
\Description[Ablation table for parameter selection components]{
The table reports R precision and dual multi-modal distance for fixed parameters, ParamGate variants with layer, step, and token selection, and test-time scaling. Performance improves as prompt-tailored components are added, with test-time scaling performing best overall.}
{\small
\begin{tabular}{lcccc}
\toprule
\multirow{2}{*}{Configuration} &
\multicolumn{3}{c}{R Precision $\uparrow$} &
\multirow{2}{*}{Dual MM Dist $\downarrow$} \\
\cmidrule(lr){2-4}
 &   Top1 & Top2 & Top3 &\\
\midrule
Fixed parameters          & 0.08& 0.16& 0.19 & 105.97\\
\schemename               &  &  &  &  \\
\quad + Layer $\ell$          & 0.08& 0.14& 0.19& 100.02\\
\quad + Step $\hat{t}$ & 0.08&   \underline{0.19}&  0.22& 99.38\\
\quad + Token $m$ &  &  &      &      \\
\quad\quad Action-detail & \textbf{0.14}&  \underline{0.19}&   \underline{0.27} & \underline{96.08} \\
\quad\quad Test-time Scaling & \underline{0.11}&  \textbf{0.30}&  \textbf{0.32} & \textbf{85.16} \\
    \bottomrule
\end{tabular}
} %
\end{table}

\subsection{Ablation} \label{sec:ablation}

\Cref{tab:ablation,fig:Ablation}, together with our supplementary video, reveal the contribution of  components of our tailored attention guidance (\cref{sec:param_select}).
In the upper raw of \cref{tab:ablation} we use a fixed set of parameters, selected according to the best performing combination in $\Theta$.
This selection yields poor results, indicating that attention modulation in the motion domain is not robust to a single configuration.

Introducing prompt-tailored parameter selection via \schemename, we incrementally add selection schemes for each parameter, resulting in progressively improved performance.
Selecting the transformer layer $\ell$ improves alignment, supporting our design choice to restrict guidance to specific layers, while incorporating adaptive diffusion-step range $\hat{t}$ further reduces deviation. Finally, for token selection, using identified action-detail tokens improves performance over fixed tokens, and test-time scaling achieves the strongest results at the cost of additional computation.

\section{Conclusion, Limitations, and Future Work}
\label{sec:Conclusions}

We presented \algoname, an unpaired, inference-time framework for text-to-motion generation from composite prompts involving simultaneous actions. 
By strengthening cross-attention scores associated with underrepresented textual components, \algoname mitigates vanishing semantics without retraining the underlying motion model. 
Observing that effective attention modulation in the motion domain is highly prompt-dependent, we introduced \schemename, a lightweight scheme that automatically selects prompt-tailored guidance parameters. 
Extensive quantitative, qualitative, and user-study evaluations demonstrate that \algoname improves semantic coverage of composite prompts while preserving motion realism, outperforming both general-purpose and composite-specific baselines.

Despite its effectiveness, MultiAct has several limitations. 
While \schemename substantially reduces manual tuning, its token-selection strategy relies on time-consuming test-time exploration or heuristic linguistic cues, incurring additional computational cost or potentially missing the most influential semantic element, respectively.
Additionally, our method depends on the qualitative performance of its backbone and therefore inherits its limitations, such as motion artifacts, excessive smoothing or jitter.
Future directions include extending our work to support more complex descriptions involving multiple interacting semantic elements. We plan to explore this by developing more expressive parameter-selection strategies, extending guidance to multiple tokens, and investigating associations between specific tokens and specific transformer layers.

\begin{acks}
We thank Theodoros Kyriakou for helping generate motions and Hila Chefer
for fruitful discussions.
This research was supported in part by the Israel Science Foundation
(grants no. 2492/20 and 1473/24) and by Len Blavatnik and the Blavatnik
Family Foundation.
This research also received funding from the EU Commission's Horizon Europe
program (grant no. 101178362).
\end{acks}

\bibliographystyle{ACM-Reference-Format}
\bibliography{Bibliography}

\ifappendix
\newpage
\appendix

\section*{\ifappendix Appendix\else Supplementary Material\fi}

This \ifappendix appendix\else supplementary material\fi{} adds details on top of the ones given in the main paper. While the main paper stands on its own, the details given here may shed more light. 

In \cref{sec:prelim_supp} we provide more details about the preliminaries of our work: motion representation, and the models MDM and DDPM.
In \cref{sec:exp_supp}, we provide 
a visualization of the user study.

\todo{if more sections are added to the appendix, describe them above. Otherwise, rephrase to "We present one section, in which ..."}

\section{Preliminaries - More Details} \label{sec:prelim_supp}

\subsection{Motion Representation} \label{sec:represent_sup}
Recall that $N$ denotes the number of frames in a motion sequence, $F$ denotes the length of the features describing a single frame, and $X\in \R^{N\times F}$ denotes a motion. Additionally, let $J$ be the number of skeletal joints.
Each feature is redundantly represented with the joint angles, positions, velocities, and foot contact~\cite{Guo:2022}.
Each single pose is defined by 
\[
(\dot{r}^{a}, \dot{r}^x, \dot{r}^z, r^y, j^p, j^r, j^v, c^f) \in \R^F,
\]
where $\dot{r}^{a} \in \R$ is the root angular velocity along the Y-axis. $\dot{r}^x, \dot{r}^z \in \R$ are root linear velocities on the XZ-plane, and $r^y \in \R$ is the root height. $j^p \in \R^{3(J-1)}$, $j^r \in \R^{6(J-1)}$ and $j^v \in \R^{3J}$ are the local joint positions, velocities, and rotations relative to the root, and $c^f \in \R^4$ are binary features denoting the foot contact labels for four foot joints (two for each leg).

\subsection{MDM and DDPM} \label{sec:mdm_ddpm_sup}

Our backbone, a modified version of  MDM~\cite{Tevet:2023:MDM}, uses DDPMs~\cite{ho2020denoising}. In the following paragraph, we recap the mechanism of DDPM.

An input motion $x_0$, is subjected to a Markov noise process consisting of T steps, resulting in the sequence $\{x_t\}_{t=0}^T$,  such that 
\begin{equation}
q(x_{t} | x_{t-1}) = \Ncal(\sqrt{\alpha_{t}}x_{t-1},(1-\alpha_{t})I),
\end{equation}
where $\alpha_{t} \in (0,1)$ are constant hyperparameters. When  $\alpha_{t}$ is small enough, we can approximate $x_{T} \sim \Ncal(0,I)$.

$x_0$ can be modeled via the reversed diffusion process by gradually cleaning $x_{T}$, using a generative network $p_\theta$. MDM~\cite{Tevet:2023:MDM} predicts the input motion, denoted $\hat{x}_{0}$, rather than $\epsilon_t$, such that   
$\hat{x}_0 = p_\theta(x_t, t)$. Then, the widespread diffusion loss is applied:

\begin{equation}
\Loss_\text{simple} = \E_{t \sim [1,T]}\norm{ x_0 - p_\theta(x_t, t)}_2^2.
\end{equation}

During inference, synthesis iterates from pure noise $x_T$. 
In each iteration, the denoising network $p_\theta$ predicts a clean version of the current sample $x_t$. The predicted clean sample $\hat{x}_{0}$ is then ``re-noised" to create the next sample $x_{t-1}$, repeatedly until  $t=0$.

\section{Experiments - More Details} \label{sec:exp_supp}

\subsection{User Study - Screenshot Visualization} \label{sec:user_study_screenshot}
\begin{figure*}[t]     
    \centering     
    \includegraphics[height=0.88\textheight]{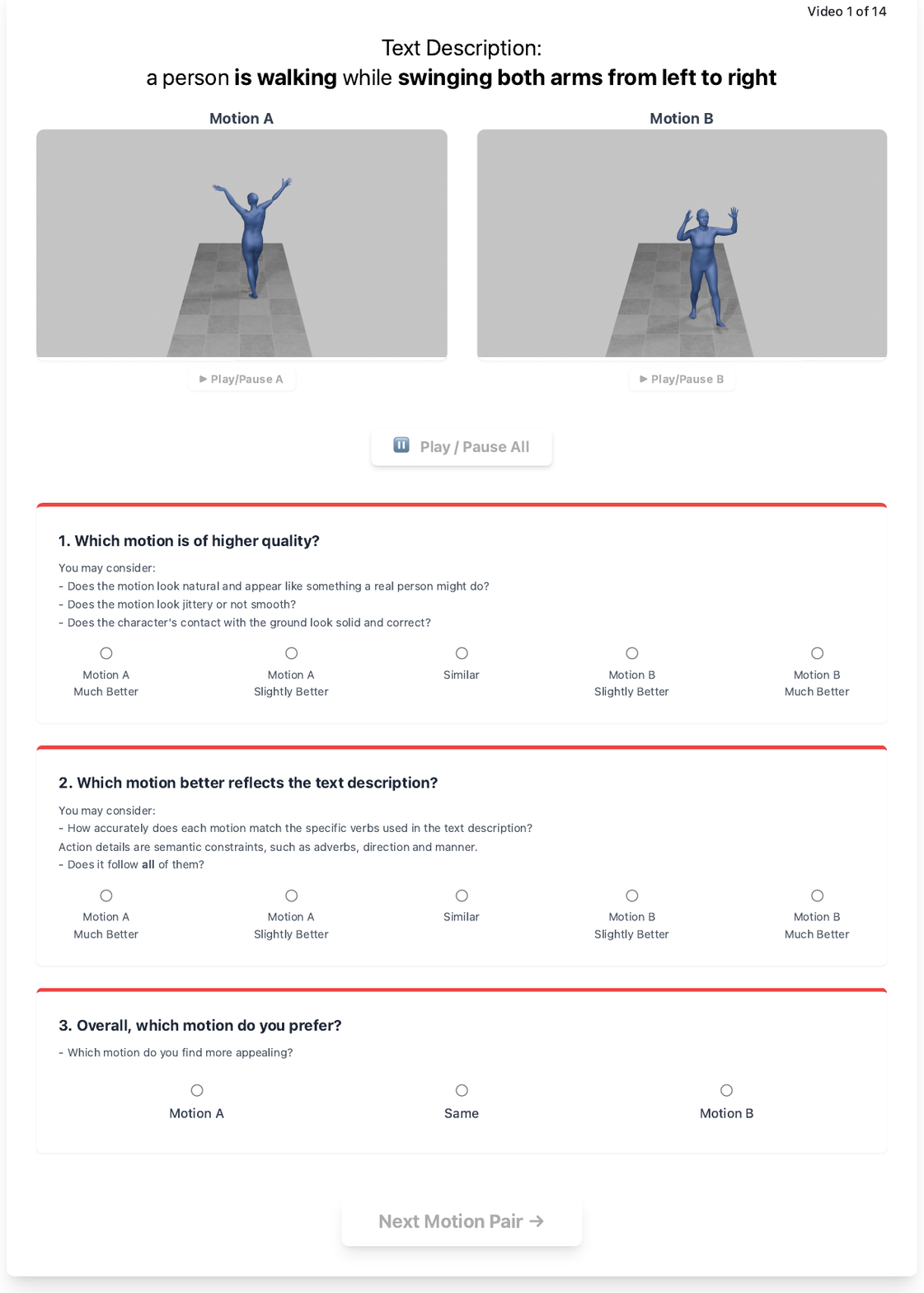} 
    \caption{\textbf{A screenshot of our user study.} Note that each human figure in the screenshot is played as a video.}     
    \label{fig:user_study_screenshot}     
    \Description[Screenshot of the user study interface]{
    The screenshot shows the interface used in the perceptual study, where participants compare animated human motions generated by different methods.}
\end{figure*}

Our user study displays a pair of video clips on each of its pages, requesting the user to select the one that is more suitable for the examined attribute, which is either quality, text alignment, or overall preference. A representative screenshot is shown in \cref{fig:user_study_screenshot}.

\fi

\end{document}